\documentclass[]{ceurart}
\usepackage{graphics}
\usepackage{xcolor}
\usepackage{svg}
\usepackage{todonotes}
\newcommand*\rot{\rotatebox{90}}

\begin{document}

\copyrightyear{2022}
\copyrightclause{Copyright for this paper by its authors.
  Use permitted under Creative Commons License Attribution 4.0
  International (CC BY 4.0).}

\conference{CLEF 2022: Conference and Labs of the Evaluation Forum, 
    September 5--8, 2022, Bologna, Italy}


\title{An End-to-End Set Transformer for User-Level Classification of Depression and Gambling Disorder}

\author[1,2]{Ana-Maria Bucur}[
orcid=0000-0003-2433-8877,
email=ana-maria.bucur@drd.unibuc.ro,
]
\address[1]{Interdisciplinary School of Doctoral Studies, University of Bucharest, Romania}
\address[2]{PRHLT Research Center, Universitat Politècnica de València, Spain}

\author[3]{Adrian Cosma}[%
orcid=0000-0003-0307-2520,
email=cosma.i.adrian@gmail.com,
]
\address[3]{Politehnica University of Bucharest, Romania}

\author[4,5]{Liviu P. Dinu}[%
orcid=0000-0002-7559-6756,
email=ldinu@fmi.unibuc.ro,
]

\address[4]{Faculty of Mathematics and Computer Science, University of Bucharest, Romania}
\address[5]{Human Language Technologies Research Center, University of Bucharest, Romania}

\author[2]{Paolo Rosso}[%
orcid=0000-0002-8922-1242,
email=prosso@dsic.upv.es,
]

\begin{abstract}
This work proposes a transformer architecture for user-level classification of gambling addiction and depression that is trainable end-to-end. As opposed to other methods that operate at the post level, we process a set of social media posts from a particular individual, to make use of the interactions between posts and eliminate label noise at the post level. We exploit the fact that, by not injecting positional encodings, multi-head attention is permutation invariant and we process randomly sampled sets of texts from a user after being encoded with a modern pretrained sentence encoder (RoBERTa / MiniLM). Moreover, our architecture is interpretable with modern feature attribution methods and allows for automatic dataset creation by identifying discriminating posts in a user's text-set. We perform ablation studies on hyper-parameters and evaluate our method for the eRisk 2022 Lab on early detection of signs of pathological gambling and early risk detection of depression. The method proposed by our team BLUE obtained the best ERDE$_5$ score of 0.015, and the second-best ERDE$_{50}$ score of 0.009 for pathological gambling detection. For the early detection of depression, we obtained the second-best ERDE$_{50}$ of 0.027.

\end{abstract}


\maketitle
\vspace{-5mm}
\section{Introduction}
How much can one know about someone from their social media interactions? Billions of people\footnote{\url{https://www.statista.com/statistics/272014/global-social-networks-ranked-by-number-of-users/}} use social media sites like Facebook, Instagram, Twitter, and Reddit every day. While some sites like Facebook and Instagram encourage users to use their real names, websites such as Reddit are often praised for enabling users to hide between a pseudonym, offering the illusion of privacy. Under the guise of anonymity, users tend to post more personal information related to their lives and their everyday struggles instead of striving to maintain an image and a persona when their identities are open \cite{de2014mental}. Many aspects of a user's personal life can be uncovered in their posting history. Of course, not one single post can be all-encompassing, but rather the information is scattered across many unrelated comments and posts. For instance, on the \textit{r/relationship\_advice}\footnote{\url{https://www.reddit.com/r/relationship_advice/}} subreddit a user might reveal their gender and age when discussing intimate relationship struggles, while on \textit{r/depression}\footnote{\url{https://www.reddit.com/r/depression/}} a user might provide clues for their internal conflicts and experiences.

In the task of mental health disorders detection from social media text, many approaches operate on the post-level \cite{tadesse2019detection,rissola2020dataset,Bucur2021EarlyRD}, considering that, for instance, if a user is depressed, then all their posts might contain some information regarding this issue. However, we posit that this method of post-level classification is unsuitable - many posts are unrelated and uninformative to the particular task. Their interaction, however, might contain clues to the mental well-being of a user. 

As such, we propose an architecture that performs user-level classification by processing a set of posts from a user. We exploit the fact that the multi-head attention operation in transformers is permutation invariant and inputs multiple texts from a single user into the network, modeling their interaction and classifying the user. This approach has several advantages: \textit{(i)} it is trainable end-to-end, mitigating the need for hand-crafted construction of global user features \textit{(ii)} it is robust to label noise, as some posts might be uninformative, the network learns to ignore them in the decision and \textit{(iii)} it is interpretable, using feature attribution methods \cite{sundararajan2017axiomatic} we can extract the most important posts for the decision.

The Early Risk Prediction on the Internet (eRisk)\footnote{\url{https://erisk.irlab.org/}} Lab started in 2017 with one pilot task and, since then, tacked the early risk detection of several mental illnesses: depression, self-harm, eating disorders, and pathological gambling. This work showcases team BLUE's proposed approach for Tasks 1 and 2 of eRisk 2022 Lab \cite{paparar2022overview}, of gambling and depression detection, respectively.

The paper makes the following contributions:
\begin{enumerate}
    \item We propose a set-based transformer architecture for user-level classification, which makes a decision by processing multiple texts of a particular user.
    \item We show that our architecture is robust to label noise and is interpretable with modern feature attribution methods, allowing it to be used as a dataset filtering tool.
    \item We obtained promising results on the eRisk 2022 tasks on early risk detection of pathological gambling (best ERDE$_5$\footnote{Early Risk Detection Error, introduced in Section \ref{sec:evaluation}} score of 0.015 and the second-best ERDE$_{50}$ score of 0.009) and depression detection (second-best ERDE$_{50}$ of 0.027).
\end{enumerate}

\section{Related Work}
\textbf{Pathological Gambling} For the detection of gambling disorder, the eRisk Lab is the first to use social media data for the assessment of gambling risk. Usually, the automated methods use data from behavioral markers \cite{philander2014identifying,deng2019applying} or personality biomarkers \cite{cerasa2018personality}. In the first iteration of the task for gambling addiction detection, the best-performing systems were developed by \citet{maupome2021early} and \citet{loyola2021unsl}. \citet{maupome2021early} used a user-level approach based on the similarity distance between the vector of topic probabilities of the users' texts to be assessed for pathological gambling risk and testimonials or items from a self-evaluation questionnaire for compulsive gamblers. By using this method, the authors obtain the best ERDE$_5$ of 0.048. \citet{loyola2021unsl} attain the best ERDE$_{50}$ (0.020) and latency-weighted F1 (0.693) through a post-level rule-based early alert policy on bag-of-words text representation classified with SVM. 

\textbf{Depression} Depression detection from social media data is an interdisciplinary topic, and efforts have been made by researchers from both NLP and Psychology to detect different markers of depression found in the online discourse of individuals. Some depression cues found in language are: greater use of the first-person singular pronouns \textit{"I"} \cite{rude2004language}, lesser use of first-person plural \textit{"we"} \cite{bucur2021psychologically}, increased use of negative or absolutist terms (e.g., "never", "forever") \cite{fekete2002internet}, greater use of verbs at past tense \cite{smirnova2018language}.

For the task of early detection of depression, the best systems from the first iteration of the task (eRisk 2017) used as input linguistic meta information extracted from the texts such as LIWC \cite{pennebaker2001linguistic}, readability and hand-crafted features \cite{trotzek2017linguistic} obtaining the best ERDE$_5$ (12.70$\%$) or a combination of linguistic information and temporal variation of terms from users' posts \cite{Errecalde2017TemporalVO} achieving the best ERDE$_{50}$ (9.68$\%$). The best-performing systems from eRisk 2018 were the ones from \citet{funez2018unsl} and \citet{trotzek2018word}. \citet{funez2018unsl} propose a user-level approach using an SVM classifier on semantic representations that take into account the temporal variation of terms between the users' posts and achieve an ERDE$_5$ of 8.78$\%$. On the other hand, the best ERDE$_{50}$ (6.44$\%$) is attained by \citet{trotzek2018word} using a chunk-level\footnote{in 2018 the test data was released in chunks of posts, not one post at a time as it is the case in this year's tasks} approach using an ensemble of logistic regression classifiers on bag-of-words features. The dataset from the depression detection task from the eRisk Lab was an important resource later used in different research articles tackling the detection problem using approaches such as a neural network architecture on topic modeling features \cite{bucur2020detecting}, SVM or deep learning architectures using fine-grained emotions features \cite{aragon2021detecting} or deep learning methods using content, writing style and emotion features \cite{uban2021emotion}.

\section{Method}
The transformer encoder, as proposed by Vaswani et al. \cite{vaswani2017attention}, essentially consists of multiple sequential layers of multi-head attention. Scaled dot-product attention of a query $Q$ relative to a set of values $V$ and a set of keys $K$ is computed using the following equation ($d_k$ is the dimensionality of the query and keys):

\begin{equation}
    \textrm{Attention}(Q, K, V) = \textrm{softmax}(\frac{QK^T}{\sqrt{d_k}}) V
\end{equation}

As such, multi-head attention consists of multiple applications of the attention mechanism to the same input. The multi-head attention is defined as:

\begin{equation}
    \begin{aligned}
    \textrm{MultiHead}(Q, K, V) = \textrm{Concat}(\textrm{head}_1, \textrm{head}_2 \dots \textrm{head}_h) W^O\\
    \textrm{head}_i = \textrm{Attention}(QW_i^Q, KW_i^K, VW_i^V)
    \end{aligned}
\end{equation}

In this formulation, multi-head attention is \textit{permutation invariant}, and the current way to inject temporal information into the input sequence is by employing positional encodings \cite{gehring2017convolutional}. This is useful when processing sequential data such as texts. However, by omitting positional encodings, the transformer essentially acts as a \textit{set encoder}. Lee et al. \cite{lee2019set} introduced the Set Transformer, in which they prove that multi-head attention is permutation invariant and that the Set Transformer is a universal approximator of permutation invariant functions. We make use of this fact to perform user-level classification by processing \textit{sets of texts} (in the form of social media posts) from a particular user. The intuition behind processing a set of texts from a user is that no single social media post is sufficiently informative for a classifier decision, but rather their interaction and the user behavior as a whole. Moreover, through mean pooling, the inevitable noise (in terms of unrelated posts) is dampened, which aids classification in weakly-supervised scenarios, such as ours, in which a user is labeled rather than all of their posts.

We consider a user $i$ to contain multiple social media posts $U_i$. A set of $K$ texts $t$ are randomly sampled from $U_i$, which defines our text-set $S_i = \{t^j \sim U_i, j \in (1\dots K)\}$. We sample $K$ posts from the user's history, instead of processing all of them due to memory limitations - some individuals have thousands of posts while others have only in the order of tens. Moreover, stochasticity is introduced in the training procedure, which prevents overfitting. As such, for training, an input batch of size $n$ is defined by the concatenation of $n$ such text-sets: $B = \{S_{b_1}, S_{b_2}, \dots S_{b_n}\}$. We do not consider the relative order of the texts for a particular user, and text-sets are fed into the transformer encoder without using positional encoding. Since some users have a total number of texts smaller than $K$, creating a batch of text-sets is impossible without padding and masking. However, to alleviate this problem, we train with an effective batch size of 1 and chose to employ gradient accumulation to simulate a larger batch size.

\begin{figure}[t!]
    \centering
    \includegraphics[width=\textwidth]{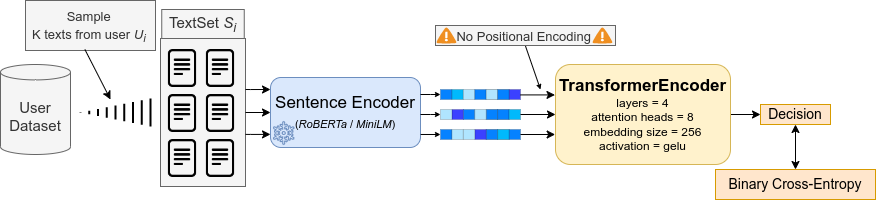}
    \caption{Proposed model architecture. We perform user-level classification by operating on a sample of K texts from a user. Texts are encoded with a pretrained sentence encoder and processed by a permutation-invariant transformer network. Binary cross-entropy loss is applied at the user level for a text-set.}
    \label{fig:architecture}
\vspace{-3mm}
\end{figure}

Figure \ref{fig:architecture} showcases our proposed model architecture for user-level classification. Each text in a text-set is embedded into a fixed-size vector using available pretrained sentence encoder models (i.e., RoBERTa / MiniLM). The text embeddings are fed into the transformer encoder network, and after processing, we perform mean pooling and output the decision. We compute binary cross-entropy at the user-level, for a text-set. The pretrained sentence encoder is frozen and not updated during training.

\begin{figure}[hbt!]
    \centering
    \includegraphics[width=\textwidth]{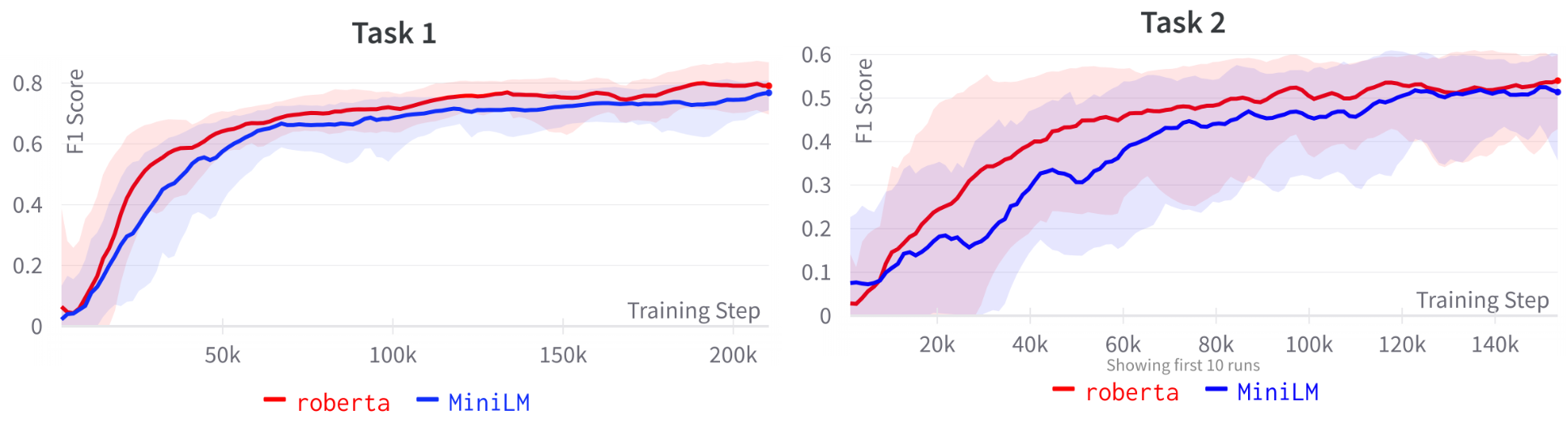}
    \caption{Performance of our model across training steps, in terms of F$_1$ score, for different sentence encoders (RoBERTa / MiniLM). We show the mean and standard deviation of F$_1$ score across multiple values of $K$. For both tasks, RoBERTa yields consistent superior performance compared to MiniLM. Best viewed in color.}
    \label{fig:encoders}
\vspace{-3mm}
\end{figure}

\begin{figure}[hbt!]
    \centering
    \includegraphics[width=\textwidth]{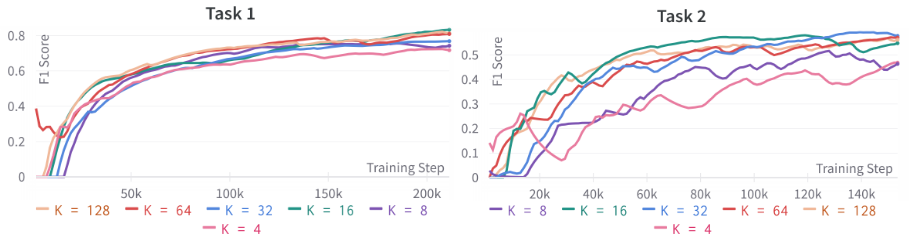}
    \caption{Performance of our model across training steps, in terms of validation F$_1$ score, for RoBERTa sentence embeddings and varying the $K$, the number of texts per user. For Tasks 1 and 2, the best performance is attained with $K=16$ and $K=32$, respectively. Best viewed in color.}
    \label{fig:num-texts}
\vspace{-5mm}
\end{figure}

\citet{baytas2017patient} proposed to use a T-LSTM to process social media posts sequentially as a time-series. The authors modify the LSTM architecture to include a relative time component. However, in our case it is unclear how to incorporate such a mechanism into the transformer architecture, aside from using a relative positional encoding \cite{kazemi2019time2vec}, which ignores long-ranged dependencies between posts. As such, we chose to ignore the temporal order of the posts and process them directly as a set. The main reason for considering the posts as a set is that in a user's post history, many posts are uninformative to the modeling task, and by processing a set of texts, label noise is reduced naturally as a direct consequence of the attention mechanism, which assigns more importance to informative posts. However, training with a sufficiently large dataset might achieve the same effect, but previous attempts at post-level classification have proven ineffective \cite{Bucur2021EarlyRD}.

In order to assess the impact of the sentence representations, we chose two different sentence encoders: RoBERTa \cite{DBLP:journals/corr/abs-1907-11692} and MiniLM \cite{wang2020minilm}. We chose RoBERTa since it is one of the best performing English language models in downstream tasks \cite{DBLP:journals/corr/abs-1907-11692}, and MiniLM, a multi-lingual model, since some users have social media posts in languages other than English. Figure \ref{fig:encoders} showcases the performance gap between the two sentence encoders, averaged across multiple values of $K$. RoBERTa yields a consistently superior performance across training steps. Similarly, to assess the impact of the text-set size $K$, we performed an ablation study, as shown in Figure \ref{fig:num-texts}. We kept the sentence encoder fixed to RoBERTa, and vary the number of texts per user $K \in \{4, 8, 16, 32, 64, 128\}$. The best performance was achieved with $K = 16$ and $K = 32$ for Tasks 1 and 2, respectively.

In our final submission, we chose RoBERTa as a sentence encoder and sampled $K = 16$ texts per user for Task 1 and $K = 32$ for Task 2. We used the standard formulation of the transformer network \cite{vaswani2017attention}, with 4 encoder layers, 8 attention heads each and a dimensionality of 256. Both networks were trained for 120 epochs, with AdamW optimizer \cite{kingma-adam}, with a cyclical learning rate \cite{smith2017cyclical} ranging from 0.00001 to 0.0001 across 6 epochs and a batch size of 128. To account for class imbalance, we computed balanced class weights with respect to each dataset and adjusted the loss function accordingly. Finally, we opted for a very high threshold when predicting the final decision.

Our proposed architecture can be easily interpretable using modern explainability methods for feature attribution \cite{lundberg2017unified,ribeiro2016should,sundararajan2017axiomatic}, such as Integrated Gradients \cite{sundararajan2017axiomatic}. It automatically identifies social media posts containing signs of mental health disorders and filters out uninformative posts.

\section{Interpretability}
Since our model operates on sets of social media texts from a particular user, we can employ model explainability methods to assess the importance of a piece of text to the model decision. Through this, automatic filtering and selection of the most indicative posts of a user can be made for use in dataset creation. This idea is similar to \citet{rissola2020dataset}, which employed a series of heuristics to recognize posts portraying depression symptoms for use in constructing a post-level training set from existing depression datasets annotated at the user level. As such, we use Integrated Gradients \cite{sundararajan2017axiomatic} to compute attribution scores for a text-set. The integrated gradients method has been used in NLP to explore the contribution of individual words and phrases to a decision made by a classifier. Since we are not operating on words, but rather on whole texts, this method computes the most important text to the classifier decision.

\begin{figure}[hbt!]
    \centering
    \includegraphics[width=\textwidth]{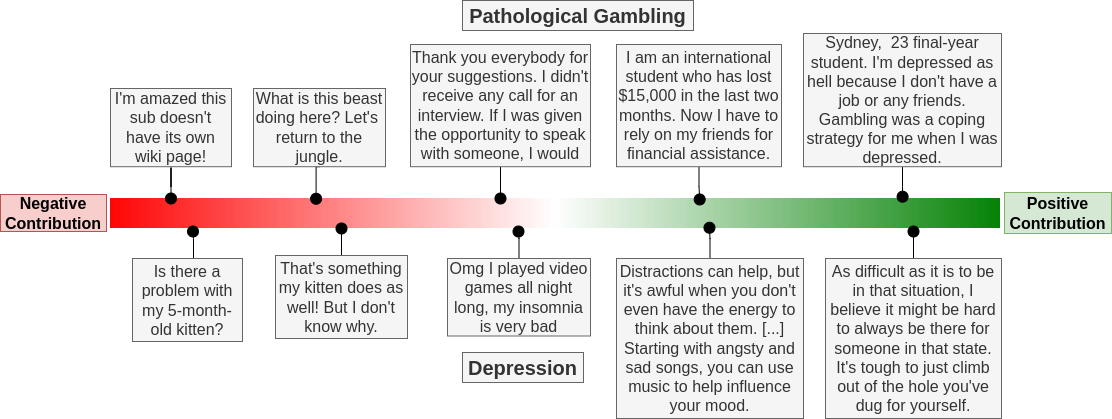}
    \caption{Texts from a particular user, relatively ranked in terms of attribution scores (contribution to a positive decision by the model) computed with the Integrated Gradients method. For each task, all texts belong to a single text-set of a user. The model is able to identify posts with a clear discriminating information for each task. Best viewed in color. Examples have been paraphrased for anonymity.}
    \label{fig:attributions}
\end{figure}

Figure \ref{fig:attributions} showcases selected samples ordered by their attribution score from the validation set of each task. All samples belong to the same user for each task, and the attribution scores are computed within the respective text-set. Posts with a high positive contribution to the decision contain more explicit descriptions of symptoms, while posts with more negative contributions are mainly unrelated to the particular mental illness. We use the integrated gradients method in one of our runs to select the most important posts in the user history. However, we emphasize that the best application of this approach is for automatic dataset creation in scenarios of weak supervision, which we aim to explore in future work.

\section{Results}
\subsection{Evaluation}
\label{sec:evaluation}

There are two kinds of evaluation used for measuring the performance of the systems, decision-based and raking-based. The \textbf{decision-based evaluation} is used for quantifying the capacity of a system to perform the binary classification and predicting if a user is from the positive class (i.e., pathological gambling or depression) or the negative one. It is comprised of standard measures for classification (Precision, Recall, F1) and measures for this specific task of early detection that consider the delay and the speed of the decision. The early risk detection error (ERDE) \cite{losada2016test} measures the correct predictions considering a late decision penalty (for predictions taken after the 5 or 50 first submissions of a user). To overcome the limitations of this metric \cite{Losada2019OverviewOE}, the \textit{latency-weighted F1} score \cite{sadeque2018measuring} was also proposed to measure the performance of early risk detection. \textit{Latency} measures the delay in detecting true positives based on the median number of submissions seen by the system before taking a decision. The \textit{speed} of a system that correctly predicts true positives from the first submission is equal to 1, while a slow system which decides after processing hundreds of texts. The latency-weighted F1 combines the F1-score with the delay in decision-taking for true positives. A perfect system should achieve a latency-weighted F1 of 1. Besides the binary classification decisions, the participating teams were asked to also submit a score for estimating the risk of users for the \textbf{ranking-based evaluation}. These scores are used to rank users' risk for pathological gambling or depression. Standard IR metrics (P@10, NDCG@10, and NDCG@100) are used to measure the models' ranking-based performance after processing 1, 100, 500, or 1000 submissions.

\subsection{Task 1: Early Detection of Signs of Pathological Gambling}
The first task proposes the detection of gambling addiction from social media data. This being the second edition of this task, the organizers provided the last year's test data for training the systems. The dataset was collected from Reddit, following the methodology described by \citet{losada2016test} and contains a chronological sequence of posts from each user. The training dataset was comprised of 164 pathological gamblers, with a total of 54,674 submissions, and 2,184 control users with 1,073,883 submissions. The test dataset contains 81 users with gambling addiction, summing 14,627 posts, and 1,998 control users with a total of 1,014,122 posts. For the testing phase, the submissions of users were released sequentially, the systems proposed by the participating teams received one submission at a time from all the users. We submitted three runs for the early detection of pathological gambling: \textbf{Run 0} is comprised of the text-set transformer model using the most recent $K = 16$ posts for prediction; the system for \textbf{Run 1} is the same text-set transformer model using as input the set of $K = 16$ texts that are most important in a user's history, selected with Integrated Gradients; \textbf{Run 2} is a baseline run, using the proposed model architecture for predicting at post-level, on one sample at a time.

\begin{table}[hbt!]
\vspace{-2mm}
    \caption{Decision-based evaluation on Task 1: Early Detection of Signs of Pathological Gambling. We show the performance of our systems compared to the best-performing run from each team.}
    \centering
    \resizebox{\textwidth}{!}{
        \begin{tabular}{lccccccccc}
        Team & Run ID & P & R & F1 & ERDE$_5$ & ERDE$_{50}$ & Latency$_{TP}$ & Speed & Latency-Weighted F1 \\
        \toprule
        BLUE & 0 & 0.260 & 0.975 & 0.410 & \textbf{0.015} & 0.009 & \textbf{1.0} & \textbf{1.000} & 0.410\\
        BLUE & 1 & 0.123 & 0.988 & 0.219 & 0.021 & 0.015 & \textbf{1.0} & \textbf{1.000} & 0.219\\
        BLUE & 2 & 0.052 & \textbf{1.000} & 0.099 & 0.037 & 0.028 & \textbf{1.0} & \textbf{1.000} & 0.099\\
        \midrule
        UNED-NLP & 4 & 0.809 & 0.938 & \textbf{0.869} & 0.020 & \textbf{0.008} & 3.0 & 0.992 & \textbf{0.862}\\
        SINAI & 1 & 0.575 & 0.802 & 0.670 & \textbf{0.015} & 0.009 & \textbf{1.0} & \textbf{1.000} & 0.670\\
        BioInfo$\_$UAVR & 4 & 0.192 & 0.988 & 0.321 & 0.033 & 0.011 & 5.0 & 0.984 & 0.316\\
        RELAI & 2 & 0.052 & 0.963 & 0.099 & 0.036 & 0.029 & \textbf{1.0} & \textbf{1.000} & 0.099\\
        BioNLP-UniBuc & 4 & 0.046 & \textbf{1.000} & 0.089 & 0.032 & 0.031 & \textbf{1.0} & \textbf{1.000} & 0.089\\
        UNSL & 1 & 0.461 & 0.938 & 0.618 & 0.041 & \textbf{0.008} & 11.0 & 0.961 & 0.594\\
        NLPGroup-IISERB & 4 & \textbf{1.000} & 0.074 & 0.138 & 0.038 & 0.037 & 41.5 & 0.843 & 0.116\\
        stezmo3 & 4 & 0.160 & 0.901 & 0.271 & 0.043 & 0.011 & 7.0 & 0.977 & 0.265\\
        \bottomrule
        \end{tabular}
    }

    \label{tab:task1-results}
\vspace{-3mm}
\end{table}
\begin{table}[hbt!]
\vspace{-2mm}
    \caption{Ranking-based evaluation on Task 1: Early Detection of Signs of Pathological Gambling.}
    \centering
    \resizebox{\textwidth}{!}{
        \begin{tabular}{lc|ccc|ccc|ccc|ccc}
        & & \multicolumn{3}{c}{1 writing} & \multicolumn{3}{c}{100 writings} & \multicolumn{3}{c}{500 writings} & \multicolumn{3}{c}{1000 writings} \\
        Team & Run ID & \rot{P@10} & \rot{NDCG@10} & \rot{NDCG@100} & \rot{P@10} & \rot{NDCG@10} & \rot{NDCG@100} & \rot{P@10} & \rot{NDCG@10} & \rot{NDCG@100} & \rot{P@10} & \rot{NDCG@10} & \rot{NDCG@100} \\
        \toprule
        BLUE & 0 & \textbf{1.00} & \textbf{1.00} & \textbf{0.76} & \textbf{1.00} & \textbf{1.00} & 0.81 & \textbf{1.00} & \textbf{1.00} & 0.89 & \textbf{1.00} & \textbf{1.00} & 0.89\\
        BLUE & 1 & \textbf{1.00} & \textbf{1.00} & \textbf{0.76} & \textbf{1.00} & \textbf{1.00} & 0.89 & \textbf{1.00} & \textbf{1.00} & 0.91 & \textbf{1.00} & \textbf{1.00} & 0.91\\
        BLUE & 2 & \textbf{1.00} & \textbf{1.00} & 0.69 & \textbf{1.00} & \textbf{1.00} & 0.40 & 0.00 & 0.00 & 0.02 & 0.00 & 0.00 & 0.01\\
        \midrule
        UNED-NLP & 4 & \textbf{1.00} & \textbf{1.00} & 0.56 & \textbf{1.00} & \textbf{1.00} & 0.88 & \textbf{1.00} & \textbf{1.00} & \textbf{0.95} & \textbf{1.00} & \textbf{1.00} & \textbf{0.95}\\
        UNSL & 0 & \textbf{1.00} & \textbf{1.00} & 0.68 & \textbf{1.00} & \textbf{1.00} & \textbf{0.90} & \textbf{1.00} & \textbf{1.00} & 0.93 & \textbf{1.00} & \textbf{1.00} & \textbf{0.95}\\
        \bottomrule
        \end{tabular}
    }
    \label{tab:task1-results-ranking}
\end{table}

Table \ref{tab:task1-results} showcases the performance of the systems measured using the decision-based measures. Regarding ERDE, our first run (Run 0), using the transformer architecture on the most recent texts from each user, manages to achieve the best ERDE$_5$ score of 0.015, and the second-best ERDE$_{50}$ score of 0.009, demonstrating that the system could detect early the true positive cases. The perfect scores for \textit{latency$_{TP}$} and \textit{speed} show that our models were successful at detecting the true positive cases after the first writing. As expected, the baseline run using a post-level approach (Run 2) has the lowest performance. Regarding Run 2, we expected it to achieve the best performance from our submitted runs, as this approach is more aggressive in taking decisions by using for classification the most informative posts from users' history. Furthermore, our best run from this year's task surpasses all the runs from our participation in the first iteration of the task in 2021 \cite{Bucur2021EarlyRD}, showing that a user-level approach considering a set of texts from each individual is more suitable than a post-level approach. In Table \ref{tab:task1-results-ranking} we show the results of the ranking-based evaluation, in which each team had to submit the rankings of users' risk for pathological gambling. Our team has excellent results for NDCG and P@10  in all the situations (after 1, 100, 1000, 5000 writings).

\subsection{Task 2: Early Detection of Depression}
This year marks the third iteration of the early detection of depression task, continuing the 2017 T1 and 2018 T2 tasks. The organizers provided the data from the previous two editions for training the models. Users from the depression class were labeled by their mention of diagnosis on their Reddit posts (e.g., "I was diagnosed with depression"). In contrast, users from the control class are users who do not have any mention of diagnosis in their posts \cite{losada2016test}. The training dataset comprises 214 users diagnosed with depression with 270,666 submissions and 1493 control users with a total of 2,959,080 submissions. The test set contains 98 users with depression with 35,332 posts, and 1,302 users in the control group with a total of 687,228 posts. The texts for making the predictions for the testing phase were released sequentially, and the systems from the participating teams had to decide on firing a decision for a specific user or waiting for more data. We submitted three runs for the early detection of depression: \textbf{Run 0} is the text-set transformer model using the most recent $K = 32$ posts for prediction; for \textbf{Run 1} we employ the same text-set transformer model using as input the set of $K = 32$ texts that are most important in a user's history, selected with Integrated Gradients; \textbf{Run 2} is a baseline run, using the proposed model architecture for predicting at post-level, on one sample at a time.

\begin{table}[hbt!]
\vspace{-2mm}
    \caption{Decision-based evaluation on Task 2: Early Detection of Depression. We show the performance of our systems compared to the best-performing run from each team.}
    \centering
    \resizebox{\textwidth}{!}{
        \begin{tabular}{lccccccccc}
        Team & Run ID & P & R & F1 & ERDE$_5$ & ERDE$_{50}$ & Latency$_{TP}$ & Speed & Latency-Weighted F1 \\
        \toprule
        BLUE & 0 & 0.395 & 0.898 & 0.548 & 0.047 & 0.027 & 5.0 & 0.984 & 0.540\\
        BLUE & 1 & 0.213 & 0.939 & 0.347 & 0.054 & 0.033 & 4.5 & 0.986 & 0.342\\
        BLUE & 2 & 0.106 & \textbf{1.000} & 0.192 & 0.074 & 0.048 & 4.0 & 0.988 & 0.190\\
        \midrule
        CYUT & 0 & 0.165 & 0.918 & 0.280 & 0.053 & 0.032 & 3.0 & 0.992 & 0.277\\
        LauSAn & 4 & 0.201 & 0.724 & 0.315 & \textbf{0.039} & 0.033 & \textbf{1.0} & \textbf{1.000} & 0.315\\
        BioInfo$\_$UAVR & 4 & 0.378 & 0.857 & 0.525 & 0.069 & 0.031 & 16.0 & 0.942 & 0.494\\
        TUA1 & 4 & 0.159 & 0.959 & 0.272 & 0.052 & 0.036 & 3.0 & 0.992 & 0.270\\
        NLPGroup-IISERB & 0 & 0.682 & 0.745 & \textbf{0.712} & 0.055 & 0.032 & 9.0 & 0.969 & \textbf{0.690}\\
        RELAI & 0 & 0.085 & 0.847 & 0.155 & 0.114 & 0.092 & 51.0 & 0.807 & 0.125\\
        UNED-MED & 1 & 0.139 & 0.980 & 0.244 & 0.079 & 0.046 & 13.0 & 0.953 & 0.233\\
        Sunday-Rocker2 & 1 & 0.355 & 0.786 & 0.489 & 0.068 & 0.041 & 27.0 & 0.899 & 0.439\\
        SCIR2 & 3 & 0.316 & 0.847 & 0.460 & 0.079 & \textbf{0.026} & 44.0 & 0.834 & 0.383\\
        UNSL & 2 & 0.400 & 0.755 & 0.523 & 0.045 & 0.026 & 3.0 & 0.992 & 0.519\\
        E8-IJS & 0 & \textbf{0.684} & 0.133 & 0.222 & 0.061 & 0.061 & 1.0 & 1.000 & 0.222\\
        NITK-NLP2 & 3 & 0.149 & 0.724 & 0.248 & 0.049 & 0.039 & 2.0 & 0.996 & 0.247\\
        \bottomrule
        \end{tabular}
    }
    \label{tab:task2-results}
\end{table}

In Table \ref{tab:task2-results} we present the performance of the systems using the decision-based metrics. Our best performing run is the transformer architecture using the most recent texts from users (Run 0), followed by the system that considers only the most informative submissions from each user for the model's decisions (Run 1). The post-level system (Run 2) has the worst performance. Our three submitted runs achieve high Recall at the expense of lower Precision scores. The precision of our models can be improved by incorporating a mechanism for weighting user posts according to the prevalence of signs of depression \cite{rissola-signs}. As such, a text-set containing few posts with signs of depression will not induce a positive prediction. Regarding the early detection evaluation, our team has the second-best score on the ERDE$_{50}$ metric (0.027), while our ERDE$_{5}$ score is close to the best one. Compared to the best metrics from the 2018 edition of this task, when the best ERDE$_{5}$ and ERDE$_{50}$ were 0.087 and 0.064, respectively, current systems surpass these scores due to more data being available for training the models and the advancements in the field of machine learning in the last few years. Regarding the standard metrics for classification, a slight improvement was made in terms of F1 score, from 0.64 in 2018 to 0.71 in 2022. The ranking-based evaluation performance from Table \ref{tab:task2-results-ranking} shows that for 1 and 1000 writings, our systems attain some of the best scores for P@10 and NDCG. 

\begin{table}[hbt!]
\vspace{-2mm}
    \caption{Ranking-based evaluation on Task 2: Early Detection of Depression.}
    \centering
    \resizebox{\textwidth}{!}{
        \begin{tabular}{lc|ccc|ccc|ccc|ccc}
        & & \multicolumn{3}{c}{1 writing} & \multicolumn{3}{c}{100 writings} & \multicolumn{3}{c}{500 writings} & \multicolumn{3}{c}{1000 writings} \\
        Team & Run ID & \rot{P@10} & \rot{NDCG@10} & \rot{NDCG@100} & \rot{P@10} & \rot{NDCG@10} & \rot{NDCG@100} & \rot{P@10} & \rot{NDCG@10} & \rot{NDCG@100} & \rot{P@10} & \rot{NDCG@10} & \rot{NDCG@100} \\
        \toprule
        BLUE & 0 & \textbf{0.80} & \textbf{0.88} & \textbf{0.54} & 0.60 & 0.56 & 0.59 & 0.80 & 0.81 & 0.66 & \textbf{0.80} & 0.80 & 0.68\\
        BLUE & 1 & \textbf{0.80} & \textbf{0.88} & \textbf{0.54} & 0.70 & 0.64 & \textbf{0.67} & 0.80 & 0.84 & \textbf{0.74} & \textbf{0.80} & \textbf{0.86} & \textbf{0.72}\\
        BLUE & 2 & \textbf{0.80} & 0.75 & 0.46 & 0.40 & 0.40 & 0.30 & 0.30 & 0.35 & 0.20 & 0.30 & 0.38 & 0.16\\
        \midrule
        NLPGroup-IISERB & 0 & 0.00 & 0.00 & 0.02 & \textbf{0.90} & 0.92 & 0.30 & \textbf{0.90} & \textbf{0.92} & 0.33 & 0.00 & 0.00 & 0.00\\
        Sunday-Rocker2 & 1 & 0.70 & 0.81 & 0.39 & \textbf{0.90} & \textbf{0.93} & 0.66 & \textbf{0.90} & 0.88 & 0.65 & 0.00 & 0.00 & 0.00\\
        \bottomrule
        \end{tabular}
    }
    \label{tab:task2-results-ranking}
\vspace{-7mm}
\end{table}

\section{Conclusion}

In this work, we proposed a transformer architecture that performs user-level classification of gambling addiction and depression detection. For each individual, the transformer processes a set of texts encoded by a pretrained sentence encoder to model the interactions between posts and mitigate noise in the dataset. Our network is interpretable and allows for automatic dataset creation by filtering uninformative posts in a user's history. Our method is a promising approach, especially for social media text processing, where a user has many texts: some informative and some unrelated to the particular modeling task. However, their interaction is indicative of the mental state of the user. We attained the best ERDE$_5$ score of 0.015, and the second-best ERDE$_{50}$ score of 0.009 for pathological gambling detection. For the early detection of depression, we obtained the second-best ERDE$_{50}$ (0.027).

For future work, we aim to extend our method and construct a mechanism for encoding the relative order of a user's posts with a modified version of relative positional embeddings \cite{qu2021explore}. While we chose an approach that ignores temporal ordering and processes posts as a set, preserving order is a natural way to increase the expressive power in modeling a user's entire social media interactions, similar to architectures such as the time-aware LSTM \cite{baytas2017patient}.

\vspace{-2mm}
\begin{acknowledgments}

The work of Ana-Maria Bucur was in the framework of the research project NPRP13S-0206-200281. The work of Paolo Rosso was in the framework of the research project PROMETEO/2019/121 (DeepPattern) by the Generalitat Valenciana. The authors thank the EU-FEDER Comunitat Valenciana 2014–2020 grant IDIFEDER/2018/025.

\end{acknowledgments}

\bibliography{refs}
\end{document}